\documentclass{sig-alternate-arxiv}

\usepackage{dsfont}

\usepackage[usenames,dvipsnames]{xcolor}
\definecolor{shadecolor}{gray}{0.9}

\newcommand{\mypar}[1]{\vspace{0.1in}\noindent \textbf{#1 \,}}

\usepackage[colorlinks,linktoc=all]{hyperref}
\usepackage[all]{hypcap}
\hypersetup{citecolor=BurntOrange}
\hypersetup{linkcolor=MidnightBlue}
\hypersetup{urlcolor=MidnightBlue}

\usepackage[numbers]{natbib}
\usepackage{subfig}

\usepackage{tikz}

\newcommand{\bY}{ \boldsymbol{Y}}
\newcommand{\bA}{ \boldsymbol{A}}
\newcommand{\g}{\,\vert\,}
\newcommand{\bzero}{\boldsymbol{0}}
\newcommand{\E}{\mathbb{E}}

\usepackage{color,soul,framed}
\usepackage{algorithm}
\usepackage{algpseudocode}
\usepackage{pifont}
\usepackage{appendix}

\makeatletter
\def\@copyrightspace{\relax}
\makeatother

\newcommand{\myeqp}[1]{\hyperref[eq:#1]{Equation~\ref*{eq:#1}}}
\newcommand{\mysec}[1]{\hyperref[sec:#1]{Section~\ref*{sec:#1}}}
\newcommand{\mytable}[1]{\hyperref[tab:#1]{Table~\ref*{tab:#1}}}
\newcommand{\myfig}[1]{\hyperref[fig:#1]{Figure~\ref*{fig:#1}}}
\newcommand{\myappendix}[1]{\hyperref[appendix:#1]{Appendix~\ref*{appendix:#1}}}
\newcommand{\myalg}[1]{\hyperref[alg:#1]{Algorithm~\ref*{alg:#1}}}

\begin{document}

\title{Modeling User Exposure in Recommendation}

\numberofauthors{4} \author{
\alignauthor
Dawen Liang\\
              \affaddr{Columbia University}\\
       \affaddr{New York, NY}\\
       \email{dliang@ee.columbia.edu}
\alignauthor
Laurent Charlin\\
             \affaddr{McGill University}\\
       \affaddr{Montreal, Canada}\\
       \email{lcharlin@cs.mcgill.ca}
\and
\alignauthor
James McInerney\\
       \affaddr{Columbia University}\\
       \affaddr{New York, NY}\\
       \email{james@cs.columbia.edu}
\alignauthor 
David M. Blei\\
       \affaddr{Columbia University}\\
       \affaddr{New York, NY}\\
       \email{david.blei@columbia.edu}
}

\maketitle
\begin{abstract}
Collaborative filtering analyzes user preferences for items (e.g., books,
movies, restaurants, academic papers) by exploiting the similarity patterns
across users. In implicit feedback settings, all the items, including the ones 
that a user did not consume, are taken into consideration. But this
assumption does not accord with the common sense understanding that users have
a limited scope and awareness of items. For example, a user might not have
heard of a certain paper, or might live too far away from a restaurant to
experience it. In the language of causal analysis \cite{imbens2015causal}, the
assignment mechanism (i.e., the items that a user is exposed to) is a latent
variable that may change for various user/item combinations. 
In this paper, we propose a new probabilistic approach that directly
incorporates \emph{user exposure} to items into collaborative filtering.
The exposure is modeled as a latent variable and the model infers its
value from data. In doing so, we recover one of the most successful
state-of-the-art approaches as a special case of our model
\cite{hu2008collaborative}, and provide a plug-in method for conditioning
exposure on various forms of \emph{exposure covariates} (e.g., topics in text,
venue locations). We show that our scalable inference algorithm
outperforms existing benchmarks in four different domains both with and
without exposure covariates.

\end{abstract}

\keywords{recommender systems, collaborative filtering, matrix factorization}

\section{Introduction}
\label{sec:introduction}

Making good recommendations is an important problem on the web. In the
recommendation problem, we observe how a set of users interacts with a
set of items, and our goal is to show each user a set of previously
unseen items that she will like.  Broadly speaking, recommendation
systems use historical data to infer users' preferences, and then use
the inferred preferences to suggest items.  Good recommendation
systems are essential as the web grows; users are overwhelmed with
choice.

Traditionally there are two modes of this problem, recommendation from
explicit data and recommendation from implicit data.  With explicit
data, users rate some items (positively, negatively, or along a
spectrum) and we aim to predict their missing ratings.  This is called
explicit data because we only need the rated items to infer a user's
preferences. Positively rated items indicate types of items that she
likes; negatively rated items indicate items that she does not like.
But, for all of its qualities, explicit data is of limited use. It is
often difficult to obtain.

The more prevalent mode is recommendation from implicit data.  In
implicit data, each user expresses a binary decision about items---for
example this can be clicking, purchasing, viewing---and we aim to
predict unclicked items that she would want to click on. Unlike
ratings data, implicit data is easily accessible.  While ratings data
requires action on the part of the users, implicit data is often a
natural byproduct of their behavior, e.g., browsing histories, click
logs, and past purchases.

But recommendation from implicit data is also more difficult than its
explicit counterpart.  The reason is that the data is binary and thus,
when inferring a user's preferences, we must use unclicked items.
Mirroring methods for explicit data, many methods treat unclicked
items as those a user does not like.  But this assumption is mistaken,
and overestimates the effect of the unclicked items.  Some of these
items---many of them, in large-scale settings---are unclicked because
the user didn't see them, rather than because she chose not to
click them.  This is the crux of the problem of analyzing implicit
data: we know users click on items they like, but we do not know why
an item is unclicked.

Existing approaches account for this by downweighting the unclicked
items.  In \citet{hu2008collaborative} the data about unclicked items are given a
lower ``confidence'', expressed through the variance of a Gaussian
random variable.  In \citet{rendle2009bpr}, the unclicked items are artificially
subsampled at a lower rate in order to reduce their influence on the
estimation.  These methods are effective, but they involve heuristic
alterations to the data.

In this paper, we take a direct approach to solving this problem.  We
develop a probabilistic model for recommendation called Exposure MF
(abbreviated as ExpoMF) that separately captures whether a user has been exposed to an item from
whether a user has ultimately decided to click on it.  This leads to an
algorithm that iterates between estimating the user preferences and
estimating the exposure, i.e., why the unclicked items were unclicked.
When estimating preferences, it naturally downweights the unclicked items
that it expected the user will like, because it imagines that she was not
exposed to them. 

Concretely, imagine a music listener with a strong preference for alternative
rock bands such as Radiohead. Imagine that, in a dataset, there are some
Radiohead tracks that this user has not listened to. There are different
reasons which may explain unlistened tracks  (e.g., the user has a limited
listening budget, a particular song is too recent or is unavailable from a
particular online service). According to that user's listening history these
unlistened tracks would likely make for good recommendations. In this situation
our model would assume that the user does not know about these tracks---she has
not been exposed to them---and downweight their (negative) contribution when
inferring that user's preferences.

Further, by separating the two sides of the problem, our approach
enables new innovations in implicit recommendation
models. Specifically, we can build models of users' exposure that
are guided by additional information such as item content, if
exposure to the items typically happens via search, or user/item
location, if the users and items are geographically organized.  

As an example imagine a recommender system for diners in New York City and
diners in Las Vegas. New Yorkers are only exposed to restaurants in New York City.
From our model's perspective, unvisited restaurants in New York are therefore more 
informative in deriving a New Yorker's preferences compared to unvisited 
restaurants in Las Vegas. Accordingly for New York users our model will
upweight unvisited restaurants in New York while downweighting unvisited
Las Vegas restaurants. 

We studied our method with user listening history from a music intelligence company, clicks from a scientific e-print server, user
bookmarks from an online reference manager, and user checkins at venues from a 
location-based social network. In all cases, ExpoMF matches or surpasses
the state-of-the-art method of \citet{hu2008collaborative}. Furthermore, when
available, we use extra information to inform our user exposure model. In those
cases using the extra information outperforms the simple ExpoMF model.
Further, when using document content information our model also
outperforms a method specially developed for recommending documents using content and user click information 
\cite{wang2011collaborative}. Finally, we illustrate the
alternative-rock-listener and the New-York-dinner examples using real data fit 
with our models in \myfig{expo_exp} and \myfig{si}.

This paper is organized as follows. We first review collaborative filtering models in \mysec{background}. We
then introduce ExpoMF in \mysec{model} and location and content
specific models in the subsequent subsection. We draw connections between ExpoMF and causal inference as well as other recommendation research paths in
\mysec{related_work}. Finally we present an empirical study in
\mysec{experiments}.

\section{Background}
\label{sec:background}

\mypar{Matrix factorization for collaborative filtering.} User-item preference
data, whether implicit or not, can be encoded in a user by item matrix.
In this paper we refer to this user by item matrix as the \emph{click
matrix} or the \emph{consumption matrix}. Given the observed entries in this matrix
the recommendation task is often framed as filling in the unobserved
entries.  Matrix factorization models, which infer (latent) user
preferences and item attributes by factorizing the click matrix, are standard in recommender
systems \cite{koren2009matrix}. From a generative modeling perspective they
can be understood as first drawing user and item latent factors corresponding,
respectively, to user preferences and item attributes. Then drawing 
observations from a specific distribution (e.g., a Poisson
or a Gaussian) with its mean parametrized by the dot product between the user and
the item factors. Formally, Gaussian matrix factorization is \cite{mnih2007probabilistic}: 
\begin{eqnarray} 	\boldsymbol\theta_{u} &\sim& \mathcal{N}(\bzero, \lambda_\theta^{-1} I_K) \nonumber \\
	\boldsymbol\beta_{i} &\sim& \mathcal{N}(\bzero, \lambda_\beta^{-1} I_K)  \nonumber \\
	y_{ui} &\sim& \mathcal{N}(\boldsymbol\theta_u^\top \boldsymbol\beta_i, \lambda_y^{-1}), \nonumber 
 \end{eqnarray}
where $\boldsymbol\theta_u$ and $\boldsymbol\beta_i$ represent user $u$'s latent preferences and
item $i$'s attributes respectively. We use the mean and covariance to
parametrize the Gaussian distribution. $\lambda_\theta$, $\lambda_\beta$, and
$\lambda_y$ are treated as hyperparameters. $I_K$ stands for the identity
matrix of dimension $K$. 

\mypar{Collaborative filtering for implicit data.} Weighted
matrix factorization (WMF), the standard factorization model for implicit
data (also known as one-class collaborative filtering \cite{pan2008one}), selectively downweights evidence from the click
matrix~\cite{hu2008collaborative}.  WMF uses a simple heuristic where all
unobserved user-item interactions are equally downweighted vis-a-vis the
observed interactions. Under WMF an observation is generated from:
\begin{eqnarray} 
y_{ui} &\sim& \mathcal{N}(\boldsymbol\theta_u^\top\boldsymbol\beta_i, c^{-1}_{y_{ui}}), \nonumber
\end{eqnarray}
where the ``confidence'' $c$ is set such that $c_1 > c_0$. This dependency between a
click and itself is unorthodox; because of it WMF is not a generative
model. As we will describe in \mysec{model} we obtain a proper generative
model by adding an exposure latent variable. 

WMF treats the collaborative filtering problem with implicit data as a
regression problem. Concretely, consumed user-item pairs are assigned a
value of one and unobserved user-item pairs are assigned a value of zero.
Bayesian personalized ranking (BPR) \cite{rendle2009bpr,
rendle2014improving} instead treats the problem as a one of ranking
consumed user-item pairs above unobserved pairs.
In a similar vein, the weighted approximate-ranking pairwise (WARP) loss
proposed in \citet{weston2011wsabie} approximately optimizes Precision@$k$. 
To deal with the non-differentiable nature of the ranking
loss, these methods typically design specific (stochastic optimization)
methods for parameter estimation.

\section{Exposure Matrix Factorization}\label{sec:model}

We present exposure matrix factorization (ExpoMF). 
In \mysec{model_description}, we describe the main model. 
In \mysec{modeling_mu} we discuss several ways 
of incorporating external information into ExpoMF (i.e., topics from text, locations). 
We derive inference procedures for our model (and variants) in \mysec{inference}. Finally we discuss how to make predictions given our model in \mysec{pred}.

\subsection{Model Description}
\label{sec:model_description}

For every combination of users $u=1,\dots,U$ and items $i=1,\dots,I$, consider two sets of variables. The first matrix $\bA = \left\{ a_{ui} \right\}$ indicates whether user $u$ has been exposed to item $i$. The second matrix $\bY = \left\{ y_{ui} \right\}$ indicates whether or not user $u$ clicked on item $i$.

\begin{figure}[!tbp]
  \centering
  \subfloat[Exposure MF.]{\includegraphics[width=0.25\textwidth]{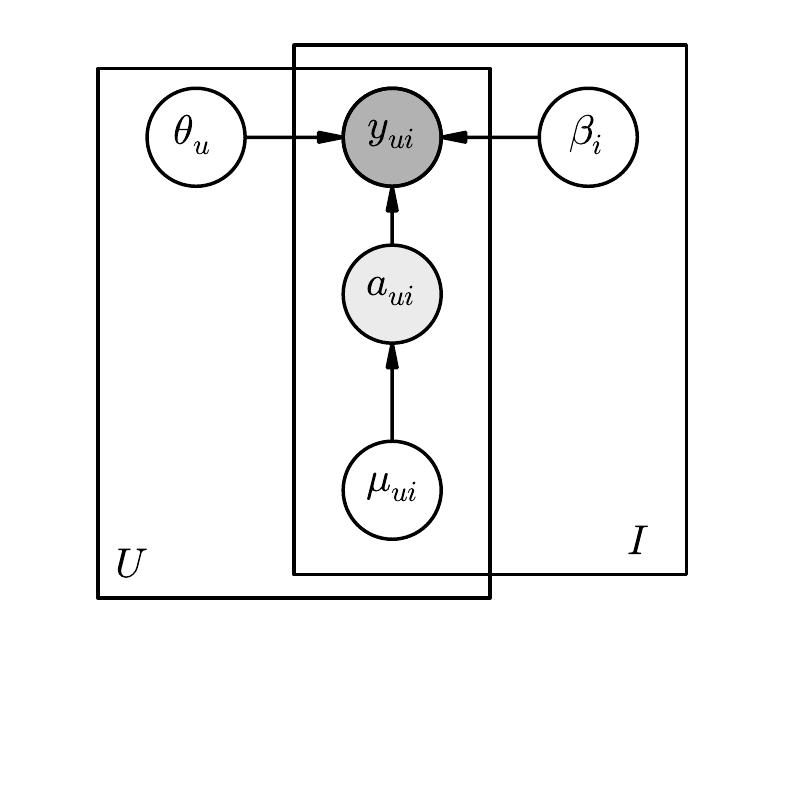}\label{fig:plate_diagram}}
    \subfloat[Exposure MF with \newline exposure covariates.]{\includegraphics[width=0.25\textwidth]{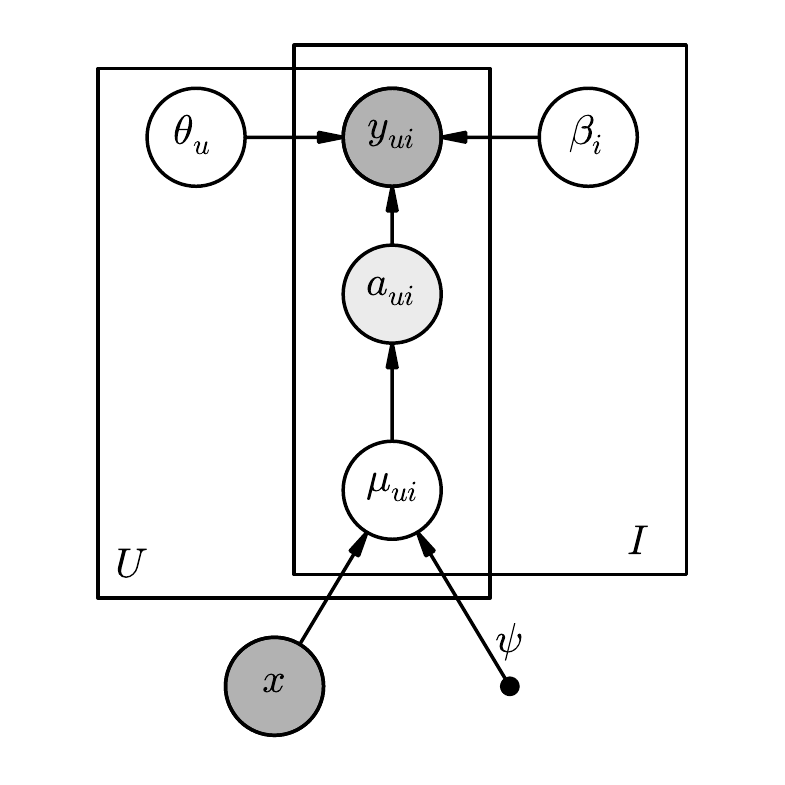}\label{fig:plate_diagram_side_info}}
  \caption{Graphical representation of the exposure MF model (both with and without exposure covariates). 
  Shaded nodes represent observed variables.
Unshaded nodes represent hidden variables. A directed
edge from node $a$ to node $b$ denotes that the variable $b$ depends
on the value of variable $a$. Plates denote replication 
by the value in the lower corner of the plate. The lightly shaded node $a_{ui}$ indicates that it is partially observed (i.e., it is observed when $y_{ui} = 1$ and unobserved otherwise).}
\end{figure}

Whether a user is exposed to an item comes from a Bernoulli. Conditional on being exposed, user's preference comes from a matrix factorization model. Similar to the standard methodology, we factorize this conditional distribution to $K$ user preferences $\theta_{i,1:K}$ and $K$ item attributes $\beta_{u,1:K}$,
\begin{eqnarray} 	\boldsymbol\theta_{u} &\sim& \mathcal{N}(\bzero, \lambda_\theta^{-1} I_K) \nonumber \\
	\boldsymbol\beta_{i} &\sim& \mathcal{N}(\bzero, \lambda_\beta^{-1} I_K)  \nonumber \\
	a_{ui} &\sim& \mathrm{Bernoulli}(\mu_{ui}) \nonumber \\
	y_{ui} \g a_{ui} = 1 &\sim& \mathcal{N}(\boldsymbol\theta_u^\top \boldsymbol\beta_i, \lambda_y^{-1}) \nonumber \\
	y_{ui} \g a_{ui} = 0 &\sim& \delta_0,
	\label{eq:gen_model}
 \end{eqnarray}
where $\delta_0$ denotes that $p(y_{ui} = 0 \g a_{ui} = 0) = 1$, and we
introduced a set of hyperparameters denoting the inverse variance
($\lambda_\theta$, $\lambda_\beta$, $\lambda_y$). $\mu_{ui}$ is the prior
probability of exposure, we discuss various ways of setting or learning it
in subsequent sections. 
A graphical representation of the model in Equation~(\ref{eq:gen_model}) is given in \myfig{plate_diagram}. 

We observe the complete click matrix $\bY$. These have a special structure. When $y_{ui} > 0$, we know that $a_{ui} = 1$. When $y_{ui} = 0$, then $a_{ui}$ is latent. 
The user might have been exposed to item $i$ and decided not to click (i.e., $a_{ui}=1$, $y_{ui} = 0$); 
or she may have never seen the item (i.e., $a_{ui}=0$, $y_{ui}=0$). 
We note that since $\bY$ is usually sparse in practice, 
most $a_{ui}$ will be latent.

The model described in Equation~(\ref{eq:gen_model})
leads to the following log joint probability\footnote{N.B., we follow the convention that $0 \log 0 = 0$ to allow the log joint to be defined when $y_{ui}>0$.} of exposures and clicks 
for user $u$ and item $i$,
\begin{flalign}
	\log p( & a_{ui}, y_{ui} \g \mu_{ui}, \boldsymbol\theta_u, \boldsymbol\beta_i, \lambda_y^{-1}) && \nonumber \\ 
	= & \log \mathrm{Bernoulli}(a_{ui} \g \mu_{ui}) + 
	a_{ui} \log \mathcal{N}(y_{ui} \g \boldsymbol\theta_u^\top \boldsymbol\beta_i, \lambda_y^{-1}) && \nonumber \\
	& + (1-a_{ui}) \log \mathbb{I}[y_{ui}=0], && 	\label{eq:a_y_joint}
\end{flalign}
where $\mathbb{I}[b]$ is the indicator function that evaluates to 1 when $b$ is true, and 0 otherwise. 

What does the distribution in \myeqp{a_y_joint} say 
about the model's exposure beliefs when no clicks are observed? 
When the predicted preference is high 
(i.e., when $\boldsymbol\theta_u^\top \boldsymbol\beta_i$ is high) 
then the log likelihood of no clicks $\log \mathcal{N}(0 \g \boldsymbol\theta_u^\top \boldsymbol\beta_i, \lambda^{-1}_y)$ is low and likely non-positive.
This feature penalizes the model 
for placing probability mass on $a_{ui}=1$, 
forcing us to believe that user $u$ is \emph{not} exposed to item $i$. 
(The converse argument also holds for low values of $\boldsymbol\theta_u^\top \boldsymbol\beta_i$). 
Interestingly, a low value of $a_{ui}$ 
downweights the evidence for $\boldsymbol\theta_u$ and $\boldsymbol\beta_i$ 
(this is clear by considering extreme values: 
when $a_{ui}=0$, the user and item factors do not affect the log joint in Eq.~\ref{eq:a_y_joint} at all; 
when $a_{ui}=1$, we recover standard matrix factorization). 
Like weighted matrix factorization (WMF) \cite{hu2008collaborative}, ExpoMF shares the same feature of selectively downweighting evidence from the click matrix. 

In ExpoMF, fixing the entries of the exposure matrix to a single value
(e.g., $a_{ui}=1, \forall u, i$) recovers Gaussian probabilistic matrix
factorization \cite{mnih2007probabilistic} (see \mysec{background}). WMF is also a special case of
our model which can be obtained by fixing ExpoMF's exposure matrix using
$c_0$ and $c_1$ as above.

The intuitions we developed for 
user exposure from the joint probability 
do not yet involve $\mu_{ui}$, 
the prior belief on exposure. 
As we noted earlier, 
there are a rich set of choices 
available in the modeling of $\mu_{ui}$. 
We discuss several of these next.

\subsection{Hierarchical Modeling of Exposure}
\label{sec:modeling_mu}

We now discuss methods for choosing and learning $\mu_{ui}$. 
One could fix $\mu_{ui}$ 
at some global value for all users and items, 
meaning that the user factors, item factors, and clicks 
would wholly determine exposure (conditioned on variance hyperparameters). 
One could also fix $\mu_{ui}$ for specific values of $u$ and $i$. 
This can be done when there is specific extra information 
that informs us about exposure (denoted as \emph{exposure covariates}), e.g. the location of a restaurant, the content of a paper. 
However, we found that empirical performance is highly sensitive to this choice, 
motivating the need to place models on the prior for $\mu_{ui}$ 
with flexible parameters. 

We introduce observed exposure covariates $\mathbf{x}_i$ and exposure model parameters $\boldsymbol\psi_{u}$ 
and condition $\mu_{ui} \g \boldsymbol{\psi}_{u}, \boldsymbol{x}_{i}$ 
according to some domain-specific structure. 
The extended graphical model with exposure covariates 
is shown in \myfig{plate_diagram_side_info}. 
Whatever this exposure model looks like, 
conditional independence between the priors for exposure 
and the more standard collaborative filtering parameters (given exposure) 
ensures that the updates for the model we introduced in \mysec{model_description}
will be the same for many popular inference procedures (e.g., expectation-maximization, variational inference, Gibbs sampling), 
making the extension to exposure covariates a plug-in procedure. 
We discuss two possible choices of exposure model next.

\mypar{Per-item $\mu_i$.} A direct way to encode exposure is via item
popularity: if a song is popular, it is more likely that you have been
exposed to it. Therefore, we choose an item-dependent conjugate prior on
$\mu_i \sim \mathrm{Beta}(\alpha_1, \alpha_2)$. This model does not use
any external information (beyond clicks). 

\mypar{Text topics or locations as exposure covariates.} 
In the domain of recommending text documents, 
we consider the exposure covariates as the set of words 
for each document. 
In the domain of location-based recommendation, 
the exposure covariates are the locations of the venues being recommended. 
We treat both in a similar way. 

Consider a $L$-dimensional ($L$ does not necessarily equal the latent space dimension $K$ in the matrix factorization model) representation $\mathbf{x}_i$ 
of the content of document $i$ 
obtained through natural language processing (e.g., word embeddings \cite{mikolov2013distributed}, latent Dirichlet allocation \cite{blei2003latent}), 
or the position of venue $i$ obtained by first clustering all the venues in the data set 
then finding the expected assignment to $L$ clusters 
for each venue. 
In both cases, $\mathbf{x}_i$ is all positive and normalizes to 1. 
Denoting $\sigma$ as the sigmoid function, 
we set
\begin{equation}
	\mu_{ui} = \sigma(\boldsymbol\psi_u^\top \mathbf{x}_i), 
\end{equation}
where we learn the coefficients $\boldsymbol\psi_u$ 
for each user $u$. Furthermore, we can include intercepts with various levels and interactions \cite{gelman2006data}. 

How to interpret the coefficients $\boldsymbol\psi_u$? 
The first interpretation is that of logistic regression, 
where the independent variables are $\mathbf{x}_i$, 
the dependent binary variables are $a_{ui}$, 
and the coefficients to learn are $\boldsymbol\psi_u$. 

The second interpretation is from a recommender systems perspective: 
$\boldsymbol\psi_u$ represents the topics (or geographical points of interest) that a user is usually exposed to, 
restricting the choice set to documents and venues that match $\boldsymbol\psi_u$. 
For example, if the $l^\mathrm{th}$ topic represents neural networks, 
and $x_{il}$ is high, 
then the user must be an avid consumer of neural network papers 
(i.e., $\boldsymbol\psi_{ul}$ must be high) 
for the model to include an academic paper $i$ in the exposure set of $u$.
In the location domain if the $l^\mathrm{th}$ cluster represents Brooklyn, 
and $x_{il}$ is high, 
then the user must live in or visit Brooklyn often 
for the model to include venues near there in the exposure set of $u$.

\subsection{Inference}
\label{sec:inference}
We use expectation-maximization (EM) \cite{dempster1977maximum} to find the maximum \emph{a posteriori} estimates of the unknown parameters of the model.\footnote{There are various other inference methods we could have used, such as Markov chain Monte Carlo \cite{smith1993bayesian} or variational inference \cite{wainwright2008graphical}. We chose EM for reasons of efficiency and simplicity, and find that it performs well in practice.} The algorithm is summarized in \myalg{exposure_mf}.

The EM inference procedure for the basic model, ExpoMF, 
can be found by writing out the full log likelihood of the model, 
then alternating between finding the expectations of missing data (exposure)
in the E(xpectation)-step and finding maximum of the likelihood with respect to the parameters in the M(aximization)-step. 
This procedure is analytical for our model because it is conditionally conjugate, 
meaning that the posterior distribution of each random variable 
is in the same family as its prior in the model. 

Furthermore, as we mentioned in \mysec{modeling_mu}, conditional independence 
between the priors for $\mu_{ui}$ and the rest of the model (given $\mu_{ui}$) 
means that the update for the latent exposure variables and user and item factors are not altered for any exposure model we use. We present these general updates first.

\mypar{E-step.}
In the E-step, we compute expectation of the exposure latent variable $\E[a_{ui}]$ for all user and item combinations ($u$, $i$) for which there are no observed clicks (recall that the presence of clicks $y_{ui}>0$ means that $a_{ui}=1$ deterministically),
\begin{equation}
\E[a_{ui} \g \boldsymbol\theta_u, \boldsymbol\beta_i, \mu_{ui}, y_{ui} = 0] = \frac{\mu_{ui} \cdot \mathcal{N}(0 | \boldsymbol\theta_u^\top\boldsymbol\beta_i, \lambda_y^{-1})}{\mu_{ui}\cdot \mathcal{N}(0 |  \boldsymbol\theta_u^\top \boldsymbol\beta_i, \lambda_y^{-1}) + (1 - \mu_{ui})}.
\label{eq:update_a}
\end{equation}
where $\mathcal{N}(0 \g \boldsymbol\theta_u^\top\boldsymbol\beta_i, \lambda_y^{-1})$ stands for the probability density function of $\mathcal{N}(\boldsymbol\theta_u^\top\boldsymbol\beta_i, \lambda_y^{-1})$ evaluated at $0$. 

\mypar{M-step.} 
For notational convenience, we define $p_{ui} = \E[a_{ui} \g \theta_u, \beta_i, \mu_{ui}, y_{ui} = 0]$ computed from the E-step. Without loss of generality, we define $p_{ui} = 1$ if $y_{ui} = 1$. The update for the latent collaborative filtering factors is:
\begin{align}
\boldsymbol\theta_u &\leftarrow \textstyle(\lambda_y \sum_i p_{ui} \boldsymbol\beta_i \boldsymbol\beta_i^\top +  \lambda_\theta I_K)^{-1} (\sum_i \lambda_y p_{ui} y_{ui} \boldsymbol\beta_i ) \label{eq:update_theta}\\
\boldsymbol\beta_i &\leftarrow \textstyle(\lambda_y \sum_u p_{ui} \boldsymbol\theta_u \boldsymbol\theta_u^\top + \lambda_\beta I_K)^{-1} (\sum_u \lambda_y p_{ui} {y}_{ui} \boldsymbol\theta_u), \label{eq:update_beta}
\end{align}

\begin{algorithm}[t!]
\caption{Inference for ExpoMF}
\label{alg:exposure_mf}
\begin{algorithmic}[1]
\State \textbf{input:} click matrix $\bY$, exposure covariates $x_{1:I}$ (topics or locations, optional).
\State \textbf{random initialization:} user factors $\boldsymbol\theta_{1:U}$, item factors $\boldsymbol\beta_{1:I}$, exposure priors~$\mu_{1:I}$ (for per-item $\mu_i$), OR exposure model parameters $\boldsymbol\psi_{1:U}$ (with exposure model).
\While {performance on validation set increases}
\State Compute expected exposure $\bA$ (\myeqp{update_a})
\State Update user factors $\boldsymbol\theta_{1:U}$ (\myeqp{update_theta})
\State Update item factors $\boldsymbol\beta_{1:I}$ (\myeqp{update_beta})
\State ExpoMF with per-item $\mu_i$:
\Statex \hspace*{.8cm} Update priors $\mu_{i}$ (\myeqp{mu_i})
\State ExpoMF with exposure model $\mu_{ui} = \sigma(\boldsymbol\psi_u^\top \mathbf{x}_i)$:
\Statex \hspace*{.8cm} Update coefficients $\boldsymbol\psi_u$ (\myeqp{update_psi} or (\ref{eq:update_psi_sgd}))
\EndWhile 
\end{algorithmic}
\end{algorithm}

\subsubsection*{Inference for the Exposure Prior $\mu_{ui}$}
\label{sec:inf_hier_model}
We now present inference for the hierarchical variants of Exposure~MF. 
In particular we highlight the updates to 
$\mu_{ui}$ under the various models we presented in \mysec{modeling_mu}.

\mypar{Update for per-item $\mu_{i}$.} Maximizing the log likelihood with respect to $\mu_i$ is equivalent to finding the mode of the complete conditional $\text{Beta}(\alpha_1 + \sum_u p_{ui}, \alpha_2 + U - \sum_u p_{ui})$, which is:
\begin{equation}\label{eq:mu_i}
\mu_i \leftarrow \frac{\alpha_1 + \sum_u p_{ui}- 1}{\alpha_1 + \alpha_2 + U - 2}
\end{equation}

\mypar{Update for exposure covariates (topics, location).} Setting $\mu_{ui} = \sigma(\boldsymbol\psi_u^\top \mathbf{x}_i)$, where $\mathbf{x}_i$ is given by pre-processing (topic analysis or clustering), presents us with the challenge of maximizing the log likelihood with respect to exposure model parameters $\boldsymbol\psi_u$. Since there is no analytical solution for the mode, we resort to following the gradients of the log likelihood with respect to $\boldsymbol\psi_u$,
\begin{equation}
	\boldsymbol\psi_u^{\text{new}} \leftarrow \boldsymbol\psi_u + \eta \nabla_{\boldsymbol\psi_u} \mathcal{L}, 	\label{eq:update_psi}
\end{equation}
for some learning rate $\eta$, where
\begin{equation}
	\nabla_{\boldsymbol\psi_u} \mathcal{L} = \textstyle \frac{1}{I}\sum_i ( p_{ui} - \sigma(\boldsymbol\psi_u^\top \mathbf{x}_i))\mathbf{x}_i.
\end{equation}

This can be computationally challenging especially for large item-set sizes $I$. Therefore, we perform (mini-batch) stochastic gradient descent: at each iteration $t$, we randomly subsample a small batch of items $B_t$ and take a noisy gradient steps:
\begin{equation} \label{eq:update_psi_sgd}
\boldsymbol\psi_u^{\text{new}} \leftarrow \boldsymbol\psi_u + \eta_t \tilde{\boldsymbol{g}}_t
\end{equation}
for some learning rate $\eta_t$, where
\begin{equation}
\tilde{\boldsymbol{g}}_t = \textstyle \frac{1}{|B_t|}\sum_{i \in B_t} (p_{ui} - \sigma(\boldsymbol\psi_u^\top \mathbf{x}_i)) \mathbf{x}_i.
\end{equation}

For each EM iteration, we found it sufficient to do a single update to the
exposure model parameter $\boldsymbol\psi_u$ (as opposed to updating until it
reaches convergence). This partial M-step~\cite{neal1998view} is much faster in
practice. 

\subsubsection*{Complexity and Implementation Details}

A naive implementation of the weighted matrix factorization (WMF) \cite{hu2008collaborative} has the same complexity as ExpoMF in terms of updating the user and item factors. However, the trick that is used to speed up computations in WMF cannot be applied to ExpoMF due to the non-uniformness of the exposure latent variable $a_{ui}$. On the other hand, the factor updates are still independent across users and items. These updates can therefore easily be parallelized.

In ExpoMF's implementation, explicitly storing the exposure matrix $\bA$ is impractical for even medium-sized datasets. As an alternative, we perform the E-step on the fly: only the necessary part of the exposure matrix $\bA$ is constructed for the updates of the user/item factors and exposure priors $\mu_{ui}$. As shown in \mysec{experiments}, with parallelization and the on-the-fly E-step, ExpoMF can be easily fit to medium-to-large datasets.\footnote{The source code to reproduce all the experimental results is available at: \url{https://github.com/dawenl/expo-mf}.}

\subsection{Prediction} \label{sec:pred}
In matrix factorization collaborative filtering the prediction of $y_{ui}$ is given by the dot product between the inferred user and item factors $\boldsymbol\theta_u^\top\boldsymbol\beta_i$. This corresponds to the predictive density of ExpoMF $p(y_{ui} \g \bY)$ using point mass approximations  to the posterior given by the EM algorithm\footnote{This quantity is also the treatment effect $\E[y_{ui} \g a_{ui}=1, \boldsymbol\theta_u, \boldsymbol\beta_i] - \E[y_{ui} \g a_{ui}=0, \boldsymbol\theta_u, \boldsymbol\beta_i]$ in the potential outcomes framework, since $a_{ui}=0$ deterministically ensures $y_{ui}=0$.}. 
However, ExpoMF can also make predictions by integrating out the uncertainty from the exposure latent variable $a_{ui}$:
\begin{equation}
\begin{split}
\E_y[y_{ui} \g \boldsymbol\theta_u, \boldsymbol\beta_i] &= \E_a\big[\E_y [y_{ui} \g \boldsymbol\theta_u, \boldsymbol\beta_i, a_{ui}]\big]\\
&= \textstyle\sum_{a_{ui}\in \{0, 1\}} \mathbb{P}(a_{ui}) \E_y [y_{ui} \g \boldsymbol\theta_u, \boldsymbol\beta_i, a_{ui}] \\
&= \mu_{ui} \cdot \boldsymbol\theta_u^\top\boldsymbol\beta_i
\end{split}
\end{equation}
We experimented with both predictions in our study and found that the
simple dot product works better for ExpoMF with per-item $\mu_i$ while
$\E[y_{ui} \g \boldsymbol\theta_u, \boldsymbol\beta_i]$ works better for ExpoMF with exposure covariates. We provide further insights about this difference in \mysec{si_doc}.

\section{Related work}
\label{sec:related_work}

In this section we highlight connections between ExpoMF and other
similar research directions.

\mypar{Causal inference.} Our work borrows ideas from the field of
causal inference~\cite{pearl2009causality,imbens2015causal}. Causal inference aims at understanding and
explaining the effect of one variable on another. 

One particular aim of causal inference is to answer counterfactual
questions. For example, ``would this new recommendation engine increase
user click through rate?''.  While online studies may answer such a question, they are
typically expensive even for large electronic commerce companies.
Obtaining answers to such questions using observational data alone (e.g.,
log data) is therefore of important practical
interest~\cite{bottou2015counterfactual,li2010contextual,swaminathan2015counterfactual}.

We establish a connection with the potential outcome framework of
\citet{rubin1974ece}. In this framework one differentiates the assignment mechanism,
whether a user is exposed to an item, from the potential outcome, whether a
user consumes an item. In potential outcome terminology our work can thus
be understood as a form a latent assignment model. In particular, while
consumption implies exposure, we do not know which items users have 
seen but not consumed. Further the questions of interest to us,
personalized recommendation, depart from traditional work in causal
inference which aims at quantifying the effect of a particular treatment
(e.g., the efficacy of a new drug).

\mypar{Biased CF models.} Authors have recognized that typical
observational data describing user rating items is biased toward items of
interest. Although this observation is somewhat orthogonal to our
investigation, models that emerged from this line of work share
commonalities with our approach.  Specifically, \citet{DBLP:conf/uai/MarlinZRS07,ling12response}
separate the \emph{selection model} (the exposure matrix) from the
\emph{data model} (the matrix factorization). However, their
interpretation, rooted in the theory of missing data~\cite{little1986statistical}, leads to a
much different interpretation of the selection model. They hypothesize
that the value of a rating influences whether or not a user will report
the rating (this implicitly captures the effect that users mostly consume
items they like \emph{a priori}). This approach is also specific to
explicit feedback data. In contrast, we model how (the value of) the
exposure matrix affects user rating or consumption. 

\mypar{Modeling exposure with random graphs.} The user-item
interaction can also be encoded as a bipartite graph.
\citet{paquet2013one} model exposure using a hidden \emph{consider graph}.
This graph plays a similar role as our exposure variable. One important
difference is that during inference, instead of directly inferring the
posterior as in ExpoMF (which is computationally more demanding), an
approximation is developed whereby a random consider graph is
stochastically sampled.

\mypar{Exposure in other contexts.} In zero-inflated Poisson regression, a
latent binary random variable, similar to our exposure variable is introduced to ``explain away'' the structural zeros, such that the
underlying Poisson model can better capture the count data
\cite{lambert1992zero}. This type of model is common in Economics where it is used
to account for overly frequent zero-valued observations.  

ExpoMF can also be considered as an instance of a spike-and-slab model~\cite{ishwaran2005spike} where the ``spike'' comes from the exposure variables and the matrix factorization component forms the flat ``slab'' part. 

\mypar{Versatile CF models.} As we show in \mysec{modeling_mu},
ExpoMF's exposure matrix can be used to model external information describing user
and item interactions. This is in contrast to most CF models which are
crafted to model a single type of data (e.g., document content when
recommendation scientific papers \cite{wang2011collaborative}). An exception is
factorization machines (FM) of \citet{rendle2010fm}. FM models all types of
(numeric) user, item or user-item features. FM considers the interaction
between all features and learns specific parameters for each interaction.

 \section{Empirical study}
\label{sec:experiments}

In this section we study the recommendation performance of ExpoMF by fitting
the model to several datasets. We provide further insights into ExpoMF's
performance by exploring the resulting model fits. We highlight that: 
\begin{itemize} 
\item ExpoMF performs comparably better than the state-of-the-art WMF
\cite{hu2008collaborative} on four datasets representing user clicks, checkins,
bookmarks and listening behavior.
\item When augmenting ExpoMF with exposure covariates its performance is
further improved. ExpoMF with location covariates and ExpoMF with content
covariates both outperform the simpler ExpoMF with per-item $\mu_i$.
Furthermore, ExpoMF with content covariates outperforms a state-of-the-art
document recommendation model \cite{wang2011collaborative}.  
\item Through posterior exploration we provide insights into ExpoMF's user-exposure modeling.
\end{itemize} 

\subsection{Datasets}
Throughout this study we use four medium to large-scale user-item consumption datasets from various domains: 
1) taste profile subset (TPS) of the million song dataset \cite{bertin2011million}; 2) scientific articles data from
arXiv\footnote{\url{http://arxiv.org}}; 3) user bookmarks from Mendeley\footnote{\url{http://mendeley.com}}; and 4) check-in
data from the Gowalla dataset \cite{cho2011friendship}. In more details:

\begin{table}
\centering
\begin{tabular}{ c c c c c  }
  \hline
   & \textbf{TPS} & \textbf{Mendeley} & \textbf{Gowalla} & \textbf{ArXiv} \\
   \hline
  \# of users & 221,830 & 45,293 & 57,629 & 37,893   \\
  \# of items & 22,781& 76,237 & 47,198 & 44,715 \\
  \# interactions & 14.0M &  2.4M & 2.3M & 2.5M\\
  $\%$ interactions & 0.29\% & 0.07\% & 0.09\% & 0.15\%\\
  \hline 
\end{tabular}
\caption{Attributes of datasets after pre-processing. Interactions are non-zero
entries (listening counts, clicks, and checkins). \% interactions refers to the
density of the user-item consumption matrix ($\bY$).}
\label{tab:data}
\end{table}

\begin{itemize}
\item \emph{Taste Profile Subset (TPS):} contains user-song play counts
collected by the music intelligence company Echo
Nest.\footnote{\url{http://the.echonest.com}} We binarize the play counts and
interpret them as implicit preference. We further pre-process the dataset by
only keeping the users with at least 20 songs in their listening history and
songs that are listened to by at least 50 users. 

\item \emph{ArXiv:} contains user-paper clicks derived from log data collected
in 2012 by the arXiv pre-print server. Multiple clicks
by the same user on a single paper are considered to be a single click. We
pre-process the data to ensure that all users and items have a minimum of 10
clicks. 

\item \emph{Mendeley:} contains user-paper bookmarks as provided by 
the Mendeley service, a ``reference
manager''. The behavior data is filtered such that each
user has at least 10 papers in her library and the papers that are bookmarked
by at least 20 users are kept. In addition this dataset contains the
content of the papers which we pre-process using standard techniques to
yield a 10K words vocabulary. In \mysec{si_doc} we make use of paper
content to inform ExpoMF's exposure model.  

\item \emph{Gowalla:} contains user-venue checkins from a location-based social
network. We pre-process the data such that all
users and venues have a minimum of 20 checkins. Furthermore, this dataset
also contains locations for the venues which we will use to guide 
location-based recommendation (\mysec{si_location}).  
\end{itemize}

The final dimensions of these datasets are summarized in \mytable{data}.

\subsection{Experimental setup} 

For each dataset we randomly split the observed user-item interactions
into training/test/validation sets with 70/20/10 proportions. In all the experiments, the dimension 
of the latent space for collaborative filtering model $K$ is 100. The
model is trained following the inference algorithm described in
\mysec{inference}. We monitor the convergence of the algorithm using
the truncated normalized discounted cumulative gain ({NDCG@100}, see below for
details) on the validation set. Hyper-parameters for ExpoMF-based models
and baseline models are also selected according to the same criterion.

To make predictions, for each user $u$, we rank each item using its predicted
preference ${y}^*_{ui} = \boldsymbol\theta_u^\top \boldsymbol\beta_i$, $i = 1, \cdots, I$.
We then exclude items from the training and validation sets and calculate all the
metrics based on the resulting ordered list. Further when using
ExpoMF with exposure covariates we found that performance was improved by predicting missing preferences
according to $\mathbb{E}[y_{ui}|\boldsymbol\theta_u,\boldsymbol\beta_i]$ (see \mysec{pred} for details).

\subsection{Performance measures}

\begin{table*}
\centering
\begin{tabular}{ c | c c | c c | c c | c c  }
   & \multicolumn{2}{c}{\textbf{TPS}} & \multicolumn{2}{c}{\textbf{Mendeley}} & \multicolumn{2}{c}{\textbf{Gowalla}} & \multicolumn{2}{c}{\textbf{ArXiv}} \\ \hline
    & WMF & ExpoMF	& WMF & ExpoMF 	& WMF& ExpoMF &	 WMF & ExpoMF \\ \hline
  Recall@20 &  0.195 &  \textbf{0.201}              & 0.128 &  \textbf{0.139}                & \textbf{0.122} & 0.118                          &  0.143 & \textbf{0.147} \\
  Recall@50 &  \textbf{0.293} &  0.286              & 0.210 & \textbf{0.221}                  & \textbf{0.192} & 0.186                         & \textbf{0.237} & 0.236 \\
  NDCG@100  &  0.255 &  \textbf{0.263}              & 0.149 & \textbf{0.159}                 & \textbf{0.118} & 0.116                         & 0.154 & \textbf{0.157} \\
  MAP@100   &  0.092 &  \textbf{0.109}             & 0.048 & \textbf{0.055}                  & \textbf{0.044} & 0.043                         & 0.051 & \textbf{0.054}\\
\end{tabular}
\caption{Comparison between WMF \cite{hu2008collaborative} and ExpoMF. While
the differences in performance are generally small, ExpoMF performs comparably better than WMF across datasets.}
\label{tab:cfresults}
\end{table*}

To evaluate the recommendation performance, we report both Recall@$k$, a standard
information retrieval measure, as well as two ranking-specific
metrics: mean average precision (MAP@$k$) and
NDCG@$k$.\footnote{\citet{hu2008collaborative} propose mean percentile
rank (MPR) as an evaluation metric for implicit feedback recommendation. Denote perc$(u, i)$ as the percentile-ranking of item $i$ within the ranked list of all items for user $u$. That is, perc$(u, i)$ = 0\% if item $i$ is ranked first in the list, and perc$(u, i)$ = 100\% if item $i$ is ranked last. Then MPR is defined as:
\begin{displaymath}
\text{MPR} = \frac{\sum_{u, i} y^{\text{test}}_{u, i} \cdot \text{perc}(u, i)}{\sum_{u, i} y^{\text{test}}_{u, i}}
\end{displaymath}
We do not use this metric because MPR penalizes ranks linearly -- if one algorithm ranks all the heldout test items in the middle of the list, while another algorithm ranks half of the heldout test items at the top while the other half at the bottom, both algorithms will get similar MPR, while clearly the second algorithm is more desirable. On the other hand, the metrics we use here will clearly prefer the second algorithm.}

We denote $\text{rank}(u, i)$ as the rank of item $i$ in user $u$'s
predicted list and $\mathbf{y}_u^{\text{test}}$ as the set of items in the
heldout test set for user $u$.

\begin{itemize}
\item Recall@$k$: For each user $u$, Recall@$k$ is computed as follows:
\begin{displaymath}
\text{Recall@}k =\sum_{i\in\mathbf{y}^{\text{test}}_u} \frac{\mathds{1}\{\text{rank}(u, i) \leq k\}}{\min(k, |\mathbf{y}_u^{\text{test}}|) }
\end{displaymath}
where $\mathds{1}\{\cdot\}$ is the indicator function. In all our experiments we report both $k=20$ and $k=50$. We do not report Precision@$k$ due to the noisy nature of the implicit feedback data: even if an item $i\notin \mathbf{y}_u^{\text{test}}$, it is possible that the user will consume it in the future. This makes Precision@$k$ less interpretable since it is prone to fluctuations. 
\item MAP@$k$: Mean average precision calculates the mean of users' average precision. The (truncated) average precision for user $u$ is: 
\begin{displaymath}
  \text{Average Precision}@k = \sum_{n=1}^k \frac{\text{Precision}@n}{\min(n,|\mathbf{y}_u^{\text{test}}|)}.
\end{displaymath}
\item NDCG@$k$: Emphasizes the importance of the top ranks by logarithmically discounting ranks. NDCG@$k$ for each user is computed as follows:
\begin{displaymath}
\text{DCG}@k = \sum_{i=1}^k \frac{2^{rel_i} - 1}{\log_2(i + 1)}; ~ \text{NDCG}@k = \frac{\text{DCG}@k}{\text{IDCG}@k}
\end{displaymath}
IDCG$@k$ is a normalization factor that ensures NDCG lies between zero and one
(perfect ranking). In the implicit feedback case the relevance is binary:
$rel_i = 1$ if $i\in \mathbf{y}_u^{\text{test}}$, and 0 otherwise. In our
study we always report the averaged NDCG across users. 
\end{itemize}

For the ranking-based measure in all the experiments we set $k=100$ which
is a reasonable number of items to consider for a user. Results are
consistent when using other values of $k$.

\subsection{Baselines}

We compare ExpoMF to weighted matrix factorization (WMF), the standard
state-of-the-art method for collaborative filtering with implicit data
\cite{hu2008collaborative}. WMF is described in \mysec{background}.

We also experimented with Bayesian personalized ranking (BPR)
\cite{rendle2009bpr}, a ranking model for implicit collaborative filtering. However preliminary results were not competitive with other
approaches. BPR is trained using stochastic optimization which can be
sensible to hyper-parameter values (especially hyper-parameters related to
the optimization procedure). A more exhaustive search over hyper-parameters
could yield more competitive results. 

We describe specific baselines relevant to modeling exposure covariates
in their dedicated subsections.

\subsection{Studying Exposure MF} 
\label{sec:expomf_study}

\begin{figure*}[!tbp]
  \centering
   \includegraphics[width=.9\textwidth]{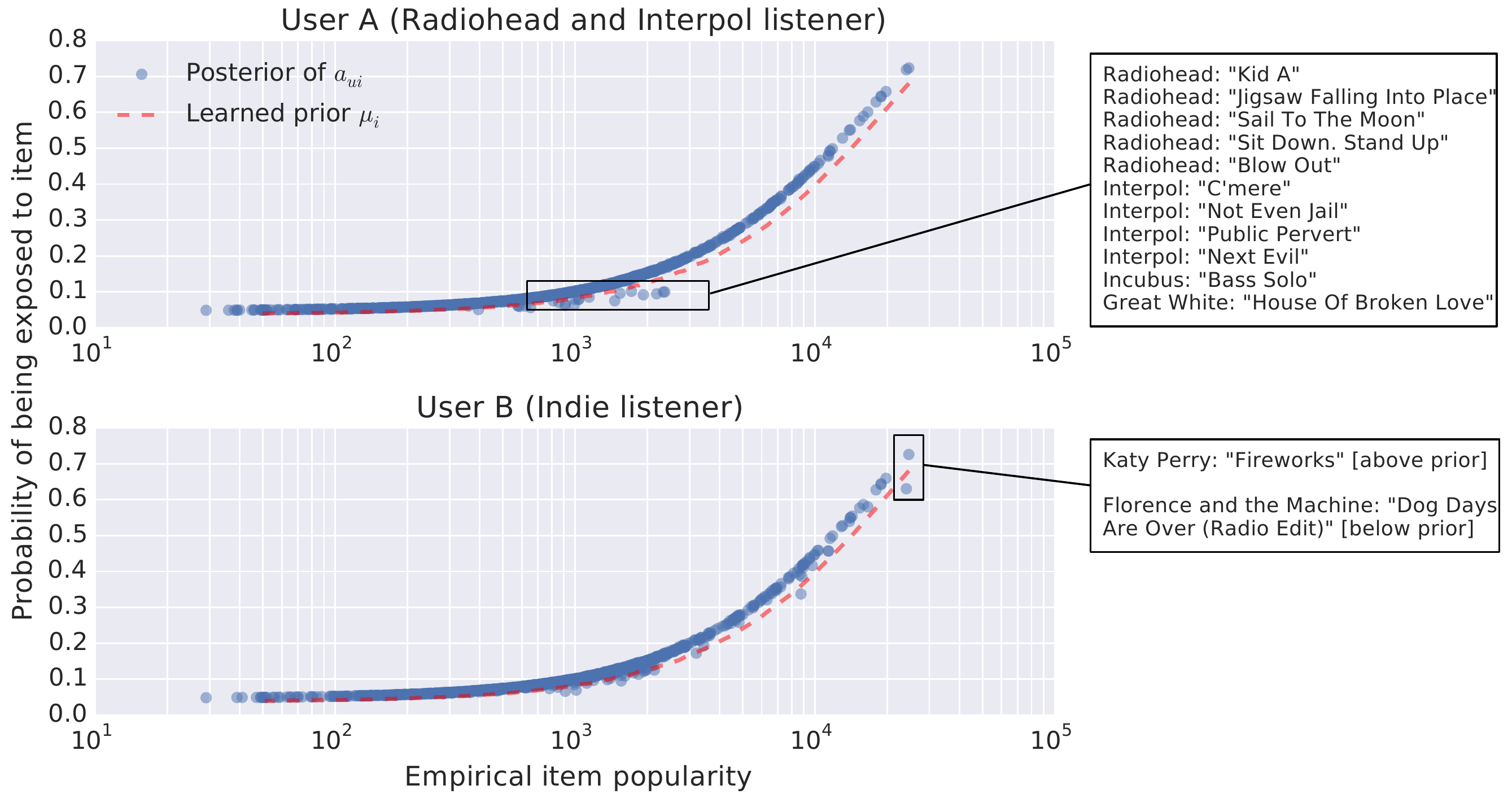}
   \caption{We compare the inferred posteriors of the exposure matrix for two
  users (denoted by blue dots) and compare against the prior probability for exposure (red dashed lined). On the top, user A is a fan of the bands Radiohead and Interpol. 
  Accordingly, the model downweights unlistened songs from these two
  bands. User B has broader interests and notably enjoys listening to
  the very popular band Florence and the Machine. Similarly as for user A,
  unlistened tracks of Florence and the Machine get downweighted in the
   posterior.}
  \label{fig:expo_exp}
\end{figure*}

\mypar{Empirical evaluation.} Results comparing ExpoMF to WMF on our four
datasets are given in
\mytable{cfresults}. Each metric is averaged across all the users.
We notice that ExpoMF performs comparably better than WMF on most
datasets (the standard errors are on the order of $10^{-4}$) though the difference in performance is small. In addition, higher values
of NDCG@100 and MAP@100 (even when Recall@$50$ is lower) indicate that the
top-ranked items by ExpoMF tend to be more relevant to users' interests.

\mypar{Exploratory analysis.} We now explore 
posterior distributions of the exposure latent variables of two specific users from the TSP dataset. This
exploration provides insights into how ExpoMF infers user exposure. 

The top figure of \myfig{expo_exp} shows the inferred exposure latent variable
$\mathbb{E}[a_{ui}]$ corresponding to $y_{ui} = 0$ for user A.
$\mathbb{E}[a_{ui}]$ is plotted along with the empirical item popularity
(measured by number of times a song was listened to in the training set).
We also plot the interpolated per-item consideration prior $\mu_i$ 
learned using \myeqp{mu_i}. There is a strong relationship between song
popularity and consideration (this is true across users). User A's
training data revealed that she has only listened to songs from either
Radiohead or Interpol (both are alternative rock bands). Therefore, for
most songs, the model infers that the probability of user A considering
them is higher than the inferred prior, i.e., it is more likely that user
A did not want to listen to them (they are \emph{true zeros}). However, as
pointed out by the rectangular box, there
are a few ``outliers'' which mostly contain songs from Radiohead and
Interpol that user A did not listen to (some of them are in fact held out
in the test set). Effectively, a lower posterior $\mathbb{E}[a_{ui}]$ than
the prior indicates that the model downweights these unlistened songs
more. In contrast, WMF downweights all songs uniformly. 

A second example is shown in the bottom figure of \myfig{expo_exp}. User B mostly
listens to \emph{indie} rock bands (e.g. Florence and the Machine, Arctic
Monkeys, and The Kills). ``Dog Days are Over'' by Florence and the Machine
is the second most popular song in this dataset, behind ``Fireworks'' by
Katy Perry. These two songs correspond to the two rightmost dots on the
figure. Given the user's listening history, the model clearly
differentiates these two similarly popular songs. The fact that user B did
not listen to ``Dog Days are Over'' (again in the test set) is more likely
due to her not having been exposed to it. In contrast the model infers
that the user probably did not like ``Fireworks'' even though it is
popular.

\subsection{Incorporating Exposure Covariates}

\myfig{expo_exp} demonstrates that ExpoMF strongly associates user
exposure to item popularity. This is partly due to the fact that the
model's prior is parametrized with a per-item term $\mu_i$. 

Here we are interested in using exposure covariates to provide additional
information about the (likely) exposure of users to items (see
\myfig{plate_diagram_side_info}). 

Recall that the role of these exposure covariates
is to allow the matrix factorization component to focus
 on items that the user has been exposed to. In particular this can be done
in the model by upweighting (increasing their probability of exposure)
items that users were (likely) exposed to and downweighting items that
were not. A motivating example with restaurant recommendations and New York
City versus Las Vegas diners was discussed in \mysec{introduction}.

In the coming subsections we compare content-aware and location-aware
versions of ExpoMF which we refer to as Content ExpoMF and Location ExpoMF
respectively. Studying each model in its respective domain we demonstrate
that the exposure covariates improve the quality of the recommendations
compared to ExpoMF with per-item $\mu_i$.

\subsubsection*{Content Covariates}\label{sec:si_doc}

Scientists---whether through a search engine, a personal recommendation
or other means---have a higher likelihood of being exposed to papers
specific to their own discipline. In this section we study the problem of
using the content of papers as a way to guide inference of the exposure
component of ExpoMF.

In this use case, we model the user exposure based on the topics of
articles. We use latent Dirichlet allocation (LDA) \cite{blei2003latent},
a model of document collections, to model article content. 
Assuming there are $K$ topics $\Phi = \boldsymbol\phi_{1:K}$, each of which is a categorical distribution over a fixed set of vocabulary, LDA treats each document as a mixture of these topics where the topic proportion $\mathbf{x}_i$ is inferred from the data. One can understand LDA as representing documents in a low-dimensional ``topic'' space with the topic proportion $\mathbf{x}_i$ being their coordinates.  

We use the topic proportion $\mathbf{x}_i$ learned from the Mendeley dataset as
exposure covariates. Following the notation of \mysec{modeling_mu}, our
hierarchical ExpoMF is:
\begin{displaymath} \mu_{ui}= \sigma(\boldsymbol\psi_u^\top \mathbf{x}_i + \gamma_u)\end{displaymath}
where we include a per-user bias term $\gamma_u$. Under this model, a
molecular biology paper and a computer science paper that a computer
scientist has not read will likely be treated differently: the model will
consider the computer scientist has been exposed to the computer science
paper, thus higher $\mathbb{E}[a_{ui}]$, yet not to the molecular biology
paper (hence lower $\mathbb{E}[a_{ui}]$). The matrix factorization
component of the model will focus on modeling computer science papers
since that are more likely to be have been exposed. 

Our model, Content ExpoMF, is trained following the algorithm in
\myalg{exposure_mf}. For updating exposure-related model parameters
$\boldsymbol\psi_{u}$ and $\gamma_u$, we take mini-batch gradient steps with a
batch-size of 10 users and a constant step size of $0.5$ for 10 epochs.

\begin{table}
\centering
\begin{tabular}{ c c c c}
\hline
            & ExpoMF & Content ExpoMF & CTR \cite{wang2011collaborative} \\ \hline
  Recall@20 & 0.139 & \textbf{0.144}         & 0.127 \\ 
  Recall@50 & 0.221 & \textbf{0.229}         & 0.210 \\ 
  NDCG@100  & 0.159 & \textbf{0.165}         & 0.150 \\ 
  MAP@100   & 0.055 & \textbf{0.056}        & 0.049 \\ 
\hline 
\end{tabular}
\caption{Comparison between Content ExpoMF and ExpoMF on Mendeley. We also compare collaborative topic regression (CTR) \cite{wang2011collaborative}, a model makes use of the same additional information as Content ExpoMF. }
\label{tab:si_doc_results}
\end{table}

\begin{figure*}
  \centering
    \includegraphics[width=\textwidth]{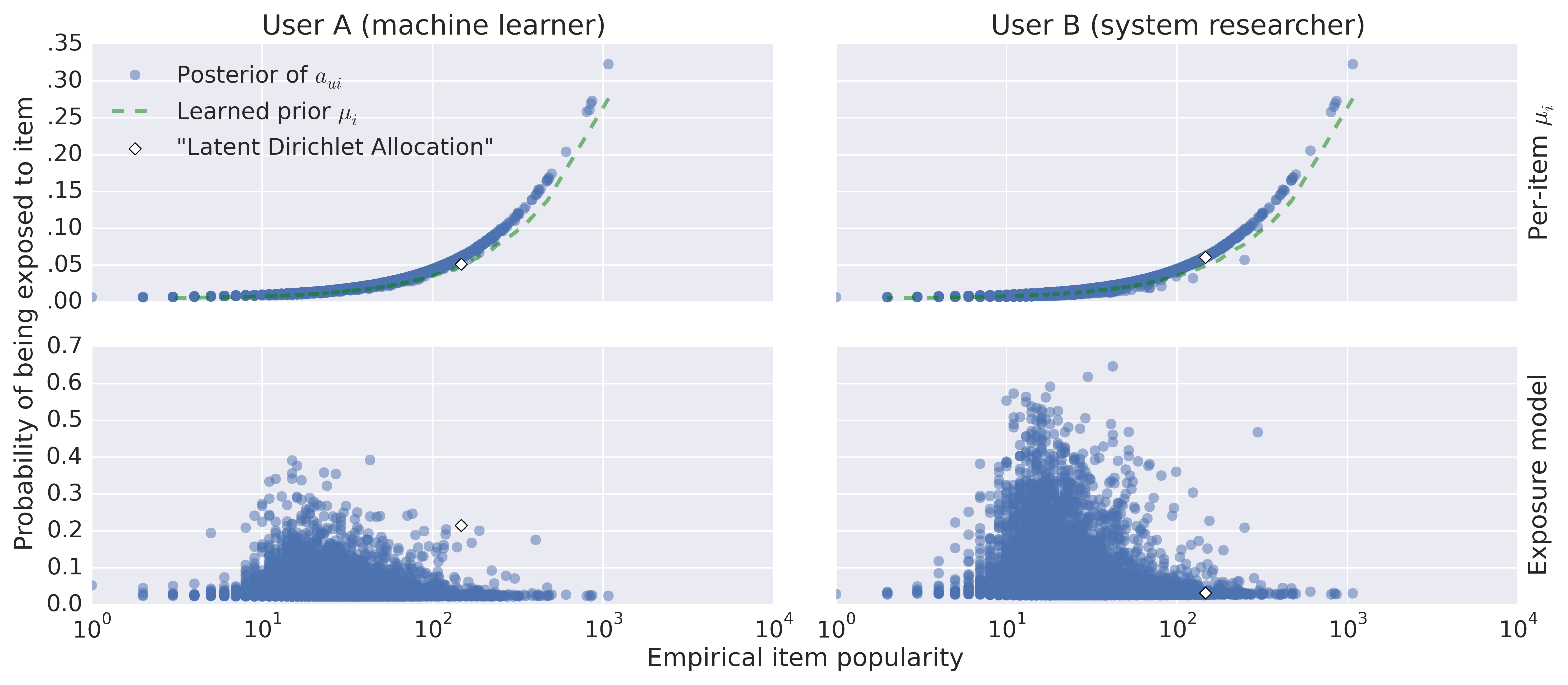}
    \caption{We compare the inferred exposure posterior of ExpoMF (top row)
    and Content ExpoMF (bottom row). On the left are the posteriors of user A
    who is interested in statistical machine learning while on the right
    user B is interested in computer system research. Neither users have
    read the ``Latent Dirichlet Allocation'' paper. ExpoMF
    infers that both users have about equal probability of having been
    exposed to it. As we discussed in \mysec{expomf_study} (and demonstrated in \myfig{expo_exp}) this is mostly based on the
    popularity of this paper. In contrast, Content ExpoMF infers that
    user A has more likely been exposed to this paper because of the closeness
    between that paper's content and user A's interest. Content ExpoMF
    therefore upweights the paper. Given user B's interests the paper is
    correctly downweighted by the model.}
    \label{fig:si}
\end{figure*}

\mypar{Study.} We evaluate the empirical performance of Content ExpoMF and report results
in \mytable{si_doc_results}. We compare to collaborative topic
regression (CTR), a state-of-the-art method for recommending scientific
papers \cite{wang2011collaborative} combining both LDA and WMF.\footnote{Note that to train CTR
we first learned a document topic model, fixed it and then learned the
user preference model. It was suggested by its authors
that this learning procedure provided computational advantages while not hindering performance significantly \citep{wang2011collaborative}.}
We did not compare with the more recent and scalable collaborative topic
Poisson factorization (CTPF) \cite{gopalan2014content} since the resulting
performance differences may have been the result of CTPF Poisson
likelihood (versus Gaussian likelihood for both ExpoMF and WMF).

We note that CTR's performance falls in-between the performance of ExpoMF
and WMF (from \mytable{data}). CTR is particularly well suited to the
\emph{cold-start} case which is not the data regime we focus on in this
study (i.e., recall that we have only kept papers that have been
bookmarked by at least 20 users).

\myfig{si} highlights the behavior of Content ExpoMF compared to that
of regular ExpoMF. Two users are selected: User A (left column) is
interested in statistical machine learning and Bayesian statistics. User B
(right column) is interested in computer systems. Neither of them have
read ``Latent Dirichlet Allocation'' (LDA) a seminal paper that falls within user A's
interests. On the top row we show the posterior of the exposure
latent variables $\mathbb{E}[a_{ui}]$ for two users (user A and user B)
inferred from ExpoMF with per-item $\mu_i$. LDA is shown using a white dot.
Overall both users' estimated exposures are dominated by the empirical item
popularity. 

In contrast, on the bottom row we plot the results of Content ExpoMF.
Allowing the model to use the documents' content to infer user exposure
offers greater flexibility compared to the simple ExpoMF model. 
This extra flexibility may also explain why there is an advantage in using
inferred exposure to predict missing observations (see \mysec{pred}).
Namely when exposure covariates are available the model can better capture the
underlying user exposures to items. In contrast using the inferred exposure to
predict with the simple ExpoMF model performs worse.

\subsubsection*{Location Covariates}
\label{sec:si_location}

When studying the Gowalla dataset we can use venue location as exposure covariates.

Recall from \mysec{modeling_mu} that location exposure covariates are created
by first clustering all venues (using $K$-means) and then finding the
representation of each venue in this clustering space. Similarly as in Content
ExpoMF (\mysec{si_doc}), Location ExpoMF departs from ExpoMF:
\begin{displaymath} \mu_{ui}= \sigma(\boldsymbol\psi_u^\top \mathbf{x}_i + \gamma_u)\end{displaymath}
where $x_{ik}$ is the venue $i$'s expected assignment to cluster $k$ and
$\gamma_u$ is a per-user bias term.\footnote{We named Content ExpoMF
and Location ExpoMF differently to make it clear to the reader that they
condition on content and location features respectively. Both models are
in fact mathematically equivalent.}

\begin{table}
\centering
\begin{tabular}{c c c c}
\hline
            & WMF & ExpoMF & Location ExpoMF \\ \hline
  Recall@20 & 0.122 & 0.118 & \textbf{0.129} \\
  Recall@50 & 0.192 & 0.186 & \textbf{0.199} \\
  NDCG@100  & 0.118 & 0.116 & \textbf{0.125} \\
  MAP@100   & 0.044 & 0.043 & \textbf{0.048} \\
\hline
\end{tabular}
\caption{Comparison between Location ExpoMF and ExpoMF with per-item
$\mu_i$ on Gowalla. Using location exposure covariates outperforms the
simpler ExpoMF and WMF according to all metrics.}
\label{tab:si_location_results}
\end{table}

\mypar{Study.} We train Location ExpoMF following the same procedure as
Content ExpoMF. We report the empirical comparison between WMF, ExpoMF and
Location ExpoMF in \mytable{si_location_results}. We note that Location
ExpoMF outperforms both WMF and the simpler version of ExpoMF. 

For comparison purposes we also developed a simple baseline FilterWMF which makes use of the
location covariates. FilterWMF filters out venues recommended by WMF that are inaccessible (too far) to the user. Since user
location is not directly available in the dataset, we estimate it using the
geometric median of all the venues the user has checked into. 
The median is preferable to the mean because it is better at handling outliers and is more likely to choose a typical visit location. 
However, the results of this simple FilterWMF baseline are worse than the results of the
regular WMF. We attribute this performance to the fact that having a single focus of location 
is too strong an assumption to capture visit behavior of users well. 
In addition, since we randomly split the data, it is possible that a user's
checkins at city A and city B are split between the training and test set.
We leave the exploration of better location-aware baselines to future work. 
 
\section{Conclusion}

In this paper, we presented a novel collaborative filtering mechanism
that takes into account user exposure to items. 
In doing so, we theoretically justify existing approaches that 
downweight unclicked items for recommendation, 
and provide an extendable framework 
for specifying more elaborate models of exposure based on logistic regression. 
In empirical studies we found 
that the additional flexibility of our model 
helps it outperform existing approaches to 
matrix factorization on four datasets from various domains. 
We note that the same approach can also be used 
to analyze explicit feedback. 

There are several promising avenues for future work. 
Consider a reader who keeps himself up to date with the ``what's new'' pages 
of a website, or 
a tourist visiting a new city looking for a restaurant recommendation. 
The exposure processes are more dynamic in these scenarios 
and may be different during training and test time. 
We therefore seek new ways to capture exposure that include 
ever more realistic assumptions about how users interact with items. 

Finally, we would like to evaluate our proposed model in a more realistic
setting, e.g., in an online environment with user interactions. It
would be instructive to evaluate the performance of ExpoMF in environments
where it may be possible to observe items which users have been exposed
to.

\section*{Acknowledgments}
This work is supported by IIS-1247664, ONR N00014-11-1-0651, DARPA FA8750-14-2-0009, Facebook, Adobe, Amazon, and the John Templeton Foundation.

{
\bibliographystyle{abbrvnat}
\bibliography{sigproc}  }
\end{document}